\documentclass[10pt,twocolumn,letterpaper]{article}

\usepackage{cvpr}              
\definecolor{cvprblue}{rgb}{0.21,0.49,0.74}
\usepackage[pagebackref,breaklinks,colorlinks,allcolors=cvprblue]{hyperref}

\usepackage[ruled,vlined]{algorithm2e}
\usepackage[accsupp]{axessibility}  

\def\paperID{xxxx} 
\def\confName{CVPR}
\def\confYear{2026}

\usepackage{multirow}
\usepackage{array}
\newcolumntype{I}{!{\vrule width 3pt}}
\newlength\savedwidth

\newlength\savewidth

\newcolumntype{x}[1]{>{\centering\arraybackslash}p{#1pt}}

\newcommand{\tablestyle}[2]{\setlength{\tabcolsep}{#1}\renewcommand{\arraystretch}{#2}\centering\footnotesize}

\title{PE3R: Perception-Efficient 3D Reconstruction}

\author{
Jie Hu, \, \, Shizun Wang, \, \,  Xinchao Wang\thanks{Corresponding author: xinchao@nus.edu.sg}\\
xML Lab, National University of Singapore
}

\begin{document}

\maketitle

\begin{abstract}
Recent advances in 2D-to-3D perception have enabled the recovery of 3D scene semantics from unposed images.
However, prevailing methods often suffer from limited generalization, reliance on per-scene optimization, and semantic inconsistencies across viewpoints.
To address these limitations, we introduce PE3R, a tuning-free framework for efficient and generalizable 3D semantic reconstruction.
By integrating multi-view geometry with 2D semantic priors in a feed-forward pipeline, PE3R achieves zero-shot generalization across diverse scenes and object categories without any scene-specific fine-tuning.
Extensive evaluations on open-vocabulary segmentation and multi-view depth estimation show that PE3R not only achieves up to 9$\times$ faster inference but also sets new state-of-the-art accuracy in both semantic and geometric metrics.
Our approach paves the way for scalable, language-driven 3D scene understanding.
Code is available at \url{github.com/hujiecpp/PE3R}.
\end{abstract}    
\section{Introduction}
\label{sec:intro}

\begin{figure*}[t]
\begin{center}
\centerline{\includegraphics[width=0.9\linewidth]{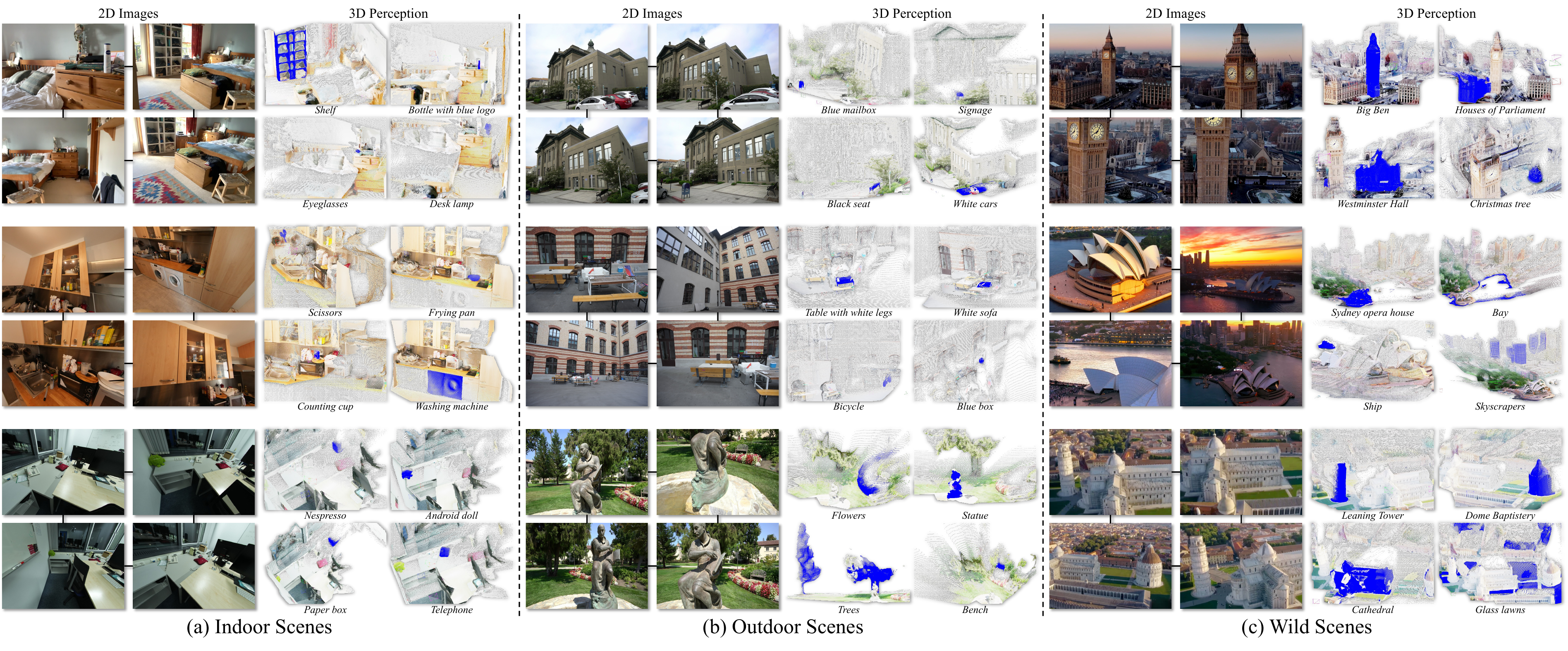}}\vspace{-10px}
\caption{\textbf{Examples of Perception-Efficient 3D Reconstruction.}
Our framework reconstructs 3D semantic scenes from unposed RGB images and supports open-vocabulary interaction, without any training or per-scene optimization.
It is efficient in two critical ways:
(1) \textit{Input efficiency}: no depth maps or camera parameters are needed;
(2) \textit{Inference efficiency}: significant speed gains on semantic reconstruction compared to prior pipelines.
These properties make PE3R practical for real-world deployment in large-scale and time-sensitive scenarios.
}
\label{fig1}
\end{center}
\vspace{-20px}
\end{figure*}

While machine vision has achieved remarkable success in 2D perception tasks, human perception is inherently three-dimensional, seamlessly integrating multiple viewpoints to construct a coherent understanding of the world~\cite{ayzenberg2024development, sinha1996role, welchman20053d}.
This fundamental gap motivates a central challenge in computer vision: can we harness powerful 2D perception models to reconstruct and comprehend 3D scenes without relying on 3D supervision?
The problem is twofold: it requires not only estimating accurate geometry from unposed images but also lifting 2D semantic information into a consistent 3D representation.

Recent methods have made significant strides in 2D-to-3D perception, recovering both geometry and semantics from unposed multi-view images~\cite{ISRF,kobayashi2022decomposing,SGISRF,tschernezki2022neural,peng2023openscene,takmaz2023openmask3d,LERF,sa3d,liu2024sanerf,guo2025wildseg3d,semanticAnyIn3dgs,gaussianGroup,feature3dgs,qin2024langsplat,SAGA}.
Despite this progress, these approaches often face a trilemma: they struggle with scene generalization, fail to maintain semantic consistency across views, and suffer from high computational costs.
For instance, Neural Radiance Fields (NeRF)~\cite{nerf} and 3D Gaussian Splatting (3DGS)~\cite{3dgs} can produce high-quality reconstructions but typically require per-scene training and involve heavy semantic lifting, making them impractical for real-time or large-scale applications.

To overcome these limitations, we introduce PE3R, a tuning-free framework for efficient, accurate, and generalizable 3D semantic reconstruction.
Inspired by recent advances in feed-forward geometry estimation~\cite{wang2024dust3r, mast3r, wang2025vggt}, PE3R employs a feed-forward pipeline that eliminates scene-specific fine-tuning while enabling rapid inference.
As illustrated in Fig.~\ref{fig1}, our framework integrates three carefully designed modules:
(1) \textit{Pixel Embedding Disambiguation} ensures semantic consistency across views and object hierarchies;
(2) \textit{Semantic Point Cloud Reconstruction} fuses geometry with dense 2D semantics to refine 3D structure;
and (3) \textit{Global View Perception} enables open-vocabulary object retrieval through language grounding.
This cohesive design empowers robust zero-shot generalization to previously unseen environments.

We conduct extensive evaluation across diverse benchmarks, including Mip-NeRF360~\cite{barron2022mip}, Replica~\cite{straub2019replica}, ScanNet++~\cite{yeshwanth2023scannet++} for open-vocabulary segmentation, and KITTI~\cite{geiger2013vision}, ScanNet~\cite{dai2017scannet}, DTU~\cite{aanaes2016large}, ETH3D~\cite{schops2017multi}, Tanks and Temples~\cite{knapitsch2017tanks} for multi-view depth estimation.
Results demonstrate that PE3R achieves up to $9\times$ faster 3D semantic reconstruction than prior methods while advancing in both semantic fidelity and geometric precision.

Our main contributions are summarized as follows:
\begin{itemize}
    \item We propose PE3R, a fast and scalable framework for 3D semantic reconstruction that operates directly on unposed 2D images, requiring no 3D supervision, per-scene fine-tuning, or camera parameters.
    \item We introduce three novel modules that address key challenges in 2D-to-3D perception: pixel embedding disambiguation for cross-view consistency, semantic point cloud reconstruction for geometry-semantics fusion, and global view perception for open-vocabulary interaction.
    \item Through comprehensive evaluations on multiple benchmarks, we demonstrate state-of-the-art performance, effective zero-shot generalization, and practical scalability.
    Code is provided to ensure reproducibility.
\end{itemize}

\section{Related Work}
\label{sec:related_work}

\noindent\textbf{2D-to-3D Reconstruction.}
Significant progress has been made in reconstructing 3D scenes from 2D images, with many methods achieving high surface fidelity and geometric coherence.
A number of approaches leverage signed distance functions (SDFs) combined with volume rendering for accurate mesh extraction~\cite{fu2022geo, guo2023streetsurf, long2022sparseneus, wang2021neus, wang2022hf}.
Neural Radiance Fields (NeRF)~\cite{nerf} popularized the use of implicit neural representations for view synthesis, while 3D Gaussian Splatting (3DGS)~\cite{3dgs} later introduced an explicit, point-based alternative that enables highly efficient rendering.
To overcome the dependency on known camera parameters, several recent works operate directly on unposed images.
Among them, DUSt3R~\cite{wang2024dust3r} predicts 3D structure from image sets.
MASt3R~\cite{mast3r} extends this direction by incorporating keypoint correspondences, and MASt3R-SfM~\cite{duisterhof2024mast3r} further unifies structure-from-motion and multi-view stereo within a transformer framework.
Hybrid techniques such as Vggsfm~\cite{wang2024vggsfm} combine traditional geometry with learned features to improve robustness.
Scalable systems like Spann3R~\cite{wang20243d}, FASt3R~\cite{Yang_2025_Fast3R}, and VGGT~\cite{wang2025vggt} integrate attention mechanisms or visibility reasoning to enhance multi-view consistency and computational efficiency.
While these methods achieve impressive geometric results, they typically lack integrated semantic understanding or require task-specific adaptation.
PE3R builds upon these geometric foundations but moves beyond them by incorporating semantic priors, enabling efficient and context-aware 3D scene understanding without per-scene training.

\begin{figure*}[t]
\begin{center}
\centerline{\includegraphics[width=0.88\linewidth]{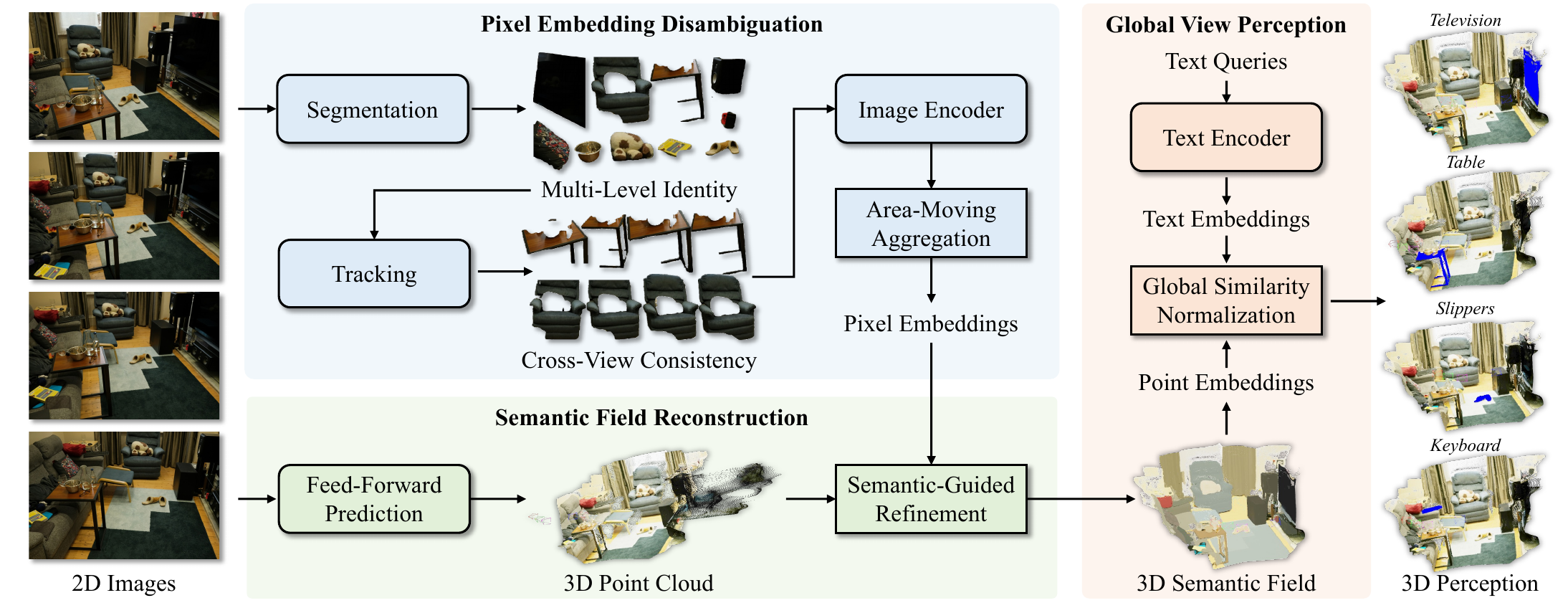}}\vspace{-10px}
\caption{\textbf{The PE3R Framework.}
Our pipeline comprises three core stages:
(1) Pixel Embedding Disambiguation: Images are decomposed into hierarchical masks (via SAM/SAM2), which are encoded by CLIP and aggregated to yield consistent per-pixel embeddings.
(2) Semantic Point Cloud Reconstruction: A feed-forward model (DUSt3R) generates 3D point clouds, which are then refined using semantic cues.
(3) Global View Perception: Text embeddings are compared against the 3D point features via global similarity normalization, enabling open-vocabulary semantic localization.
}
\label{fig2}
\end{center}
\vspace{-20px}
\end{figure*}

\noindent\textbf{2D-to-3D Perception.}
A parallel line of research seeks to lift 2D semantic information into 3D reconstructions.
Early works such as GNeRF~\cite{chen2023gnesf} and InPlace~\cite{inplace} embedded 2D masks into NeRF representations to enable 3D-aware segmentation with 2D supervision.
Subsequent efforts extended NeRF to support 3D instance segmentation, using either supervised~\cite{liu2023instance, contrastiveLift, panopticLift} or unsupervised~\cite{giraffe, yu2021unsupervised} learning paradigms.
The introduction of the Segment Anything Model (SAM)~\cite{sam} inspired a range of SAM-guided 3D lifting pipelines, including SA3D~\cite{sa3d} and Feature3DGS~\cite{feature3dgs}, which incorporate prompt-driven segmentation into 3D scene representations.
Several other methods directly project SAM masks into 3D space, such as SAGA~\cite{SAGA}, Gaussian Grouping~\cite{gaussianGroup}, SAGS~\cite{semanticAnyIn3dgs}, Click-Gaussian~\cite{choi2024click}, and FlashSplat~\cite{shen2024flashsplat}.
The advent of vision-language models has further enabled open-vocabulary 3D segmentation.
For example, LangSplat~\cite{qin2024langsplat} aligns CLIP embeddings with 3D Gaussians, while GOI~\cite{qu2024goi} enforces multi-view consistency through text-image alignment.
The Large Spatial Model (LSM)~\cite{fan2024large} performs open-vocabulary semantic segmentation by lifting features from a pre-trained 2D model (LSeg) and aggregating them into a unified 3D representation.
A fundamental limitation shared by most of these methods is their reliance on scene-specific training, iterative semantic optimization, or manual heuristics, which constrains their generalization and practical efficiency.
In contrast, PE3R supports scalable, zero-shot 3D semantic reconstruction through a cohesive feed-forward pipeline that jointly optimizes geometry and semantics, eliminating the need for 3D supervision or per-scene training.

\vspace{-0.7mm}
\noindent\textbf{2D Foundational Models.}
The advancement of large-scale pre-trained models has been pivotal for generalizable visual perception.
CLIP~\cite{radford2021learning} established a shared image-text embedding space through contrastive learning, with SigLIP~\cite{zhai2023sigmoid} offering a more efficient variant.
SAM~\cite{sam} enables promptable segmentation of arbitrary image regions, with SAM2~\cite{sam2} enhancing cross-view and temporal consistency.
DINOv2~\cite{oquab2023dinov2} delivers high-quality visual features through self-supervised learning, and Grounding DINO~\cite{liu2023grounding} facilitates open-set object detection using language-based supervision.
PE3R strategically leverages these foundational models not merely as feature extractors, but as core components within a novel pipeline designed to generate robust, language-grounded 3D semantic representations, all without requiring any 3D model fine-tuning.
\vspace{-2.1mm}
\section{Method}
\label{sec:method}

\subsection{Problem Formulation}
We address the challenging problem of reconstructing a semantically rich 3D scene representation from a collection of 2D images captured in unconstrained, in-the-wild settings.
The core difficulty stems from the absence of any geometric supervision of semantics.
To bridge this gap, we aim to reconstruct a 3D semantic point cloud directly from a set of unposed 2D images, capturing both the spatial structure and high-level semantic content of the scene.
Our framework is designed to operate without any 3D semantic annotations, camera intrinsics, or extrinsics, ensuring scalability and practicality in unconstrained environments.
The reconstructed scene supports natural language interaction, allowing users to issue open-ended text queries (\eg, \textit{``black chair'', ``white bottle''}) to retrieve and localize corresponding semantic entities in 3D, thereby enabling interactive and human-aligned perception.

\subsection{PE3R Framework} 
As illustrated in Fig.~\ref{fig2}, the PE3R pipeline is architected around three cohesive stages that collectively address the challenge of 3D semantic reconstruction from unposed images.
(1) Pixel Embedding Disambiguation:
This stage focuses on deriving view-consistent and semantically coherent embeddings from the input images.
Each image is first decomposed into hierarchical object masks using SAM~\cite{sam} or SAM2~\cite{sam2}.
These masks are encoded by a vision-language model (\eg, CLIP~\cite{radford2021learning}) into a shared semantic space.
The core of this module lies in an area-weighted spherical interpolation strategy that aggregates region features across different views and hierarchical levels.
This process resolves semantic ambiguities caused by occlusions or nested objects, yielding dense, per-pixel semantic embeddings that are consistent across images.
(2) Semantic Point Cloud Reconstruction: 
Building upon the disambiguated 2D embeddings, this stage transitions to 3D geometric reconstruction.
A feed-forward model, such as DUSt3R~\cite{wang2024dust3r}, is leveraged to generate an initial dense 3D point cloud from the multi-view images.
The key innovation here is the semantic-guided refinement of this raw geometry.
The point cloud is refined by leveraging the previously computed semantic embeddings to filter spatial outliers and enhance coherence, resulting in a unified reconstruction that is both geometrically accurate and semantically labeled.
(3) Global View Perception:
The final stage enables open-vocabulary interaction with the reconstructed 3D scene.
A user's natural language query is first encoded into a textual embedding using CLIP's text encoder.
This query embedding is then compared against the semantic features of every 3D point via cosine similarity.
To ensure a consistent and interpretable retrieval process across all views, a global min-max normalization is applied to the similarity scores.
3D points whose normalized similarity scores exceed a predefined threshold are retrieved and precisely localized, thereby facilitating real-time, language-grounded understanding and exploration of the scene.

\subsection{Pixel Embedding Disambiguation}
Establishing a consistent semantic embedding space across multiple views is foundational for robust 3D reconstruction.
This process, however, is inherently challenged by two types of ambiguity:
(i) \textit{object-level ambiguity}, where overlapping or nested structures (\eg, a cup on a table) create uncertain semantic associations;
and (ii) \textit{viewpoint-level ambiguity}, where occlusions or perspective changes lead to inconsistent semantics across different viewpoints.
To resolve these issues, we construct dense per-pixel embeddings that are both discriminative at the object level and consistent across all views.
In our formulation, an ``object'' is defined as a collection of hierarchically related masks with substantial spatial overlap.
Specifically, a smaller mask is grouped into a larger object if their Intersection over Union (IoU) exceeds a threshold of 0.9.
This simple yet effective heuristic allows us to associate fine-grained parts (\eg, chair legs) with their corresponding whole-object regions, without requiring any external supervision or predefined labels.

\noindent\textbf{Image Embedding Extraction.}
Given $n$ input images $\mathbf{X}^{1}, \ldots, \mathbf{X}^{n} \in \mathbb{R}^{3 \times H \times W}$, we employ SAM~\cite{sam} and SAM2~\cite{sam2} to decompose each image into a hierarchy of object masks with consistent object indices across views.
This yields a collection of masked image regions $\mathbf{M}^{1}, \ldots, \mathbf{M}^{n} \in \mathbb{R}^{m \times 3 \times H \times W}$, where $m$ is the number of masks per image.
A vision-language encoder $\mathcal{F}_{img}(\cdot)$ is used to extract each masked region into the semantic space:
\begin{equation}
\mathbf{F}^1,\mathbf{F}^2,\ldots,\mathbf{F}^n = \mathcal{N}(\mathcal{F}_{img}(\mathbf{M}^{1}, \mathbf{M}^{2}, \ldots, \mathbf{M}^{n})),
\end{equation}
where $\mathbf{F}^i \in \mathbb{R}^{m \times d}$ denotes the L2-normalized ($\mathcal{N}(\cdot)$) embedding matrix for the $i$-th image.

\noindent\textbf{Area-Weighted Interpolation.}
To resolve semantic ambiguities and align features across both views and hierarchy levels, we introduce a geometry-aware aggregation strategy.
Our intuition is that masks with larger visible areas typically provide more stable and reliable semantic features, and thus should exert a greater influence on the aggregated representation.
Given two unit embeddings $\mathbf{F}_A$ and $\mathbf{F}_B$ with corresponding areas $\text{area}_A$ and $\text{area}_B$, we define the area ratio $t = \frac{\text{area}_B}{\text{area}_A + \text{area}_B}$ and compute the aggregated embedding as:
\begin{equation}
\hat{\mathbf{F}}_B = a\mathbf{F}_A+b\mathbf{F}_B, a=\frac{\sin((1-t)\theta)}{\sin(\theta)}, b=\frac{\sin(t\theta)}{\sin(\theta)},
\end{equation}
where $\theta$ is the angle between $\mathbf{F}_A$ and $\mathbf{F}_B$.
This formulation possesses two key properties that underpin its effectiveness:

\textbf{Proposition 1.} \textit{Vector Normalization:} The interpolated vector $\hat{\mathbf{F}}_B$ preserves unit norm, ensuring it remains within the original semantic embedding space.

\textbf{Proposition 2.} \textit{Semantic Guidance:} If $\mathbf{F}_A$ has a higher similarity to a reference semantic vector $\mathbf{F}_C$ than $\mathbf{F}_B$ does, then $\hat{\mathbf{F}}_B$ will also exhibit a higher similarity to $\mathbf{F}_C$ than $\mathbf{F}_B$ does. This steers the aggregated representation toward more semantically meaningful directions.

These properties guarantee that our interpolation strategy is both geometrically sound and semantically informative.

\noindent\textbf{Pixel Embedding Ensemble.}
To enhance semantic coherence, we perform a two-stage ensemble that operates both within and across views.
(1) Within-view aggregation:
For each image, object masks are processed in descending order of area.
The embeddings of smaller masks (likely object parts) are iteratively aggregated via area-weighted interpolation with those of larger masks (likely whole objects), promoting semantic consistency across the hierarchical part-whole structure within a single view.
(2) Cross-view aggregation:
We seed the SAM2 tracker on the first view using SAM (segment-everything) masks; for each subsequent view, SAM2 propagates tracked masks and SAM proposes all masks again. Any SAM mask with IoU$<$0.1 to \emph{all} current SAM2 masks is treated as newly discovered/re-detected and inserted as a new tracking prompt (improving recall under tracking loss). When association is unreliable, we skip cross-view fusion and fall back to within-view disambiguation.
This two-level ensemble produces a set of stable and semantically coherent object embeddings that are consistent across both spatial hierarchies and multiple views.
The final object-level embeddings are then projected back into the pixel space, yielding dense per-pixel semantic maps $\mathbf{E}^1, \mathbf{E}^2, \ldots, \mathbf{E}^n \in \mathbb{R}^{H \times W \times d}$.
%

\begin{table}[t]
\tablestyle{6.0pt}{0.8}
\centering
\begin{tabular}{c|c|cccc}
\toprule
Dataset & Method & mIoU & mPA & mP \\
\midrule
\multirow{5}{*}{Mip.} 
&LERF~\cite{LERF}                 & 0.2698 & 0.8183 & 0.6553 \\
&F-3DGS~\cite{feature3dgs}  & 0.3889 & 0.8279 & 0.7085 \\
&GS Grouping~\cite{gaussianGroup} & 0.4410 & 0.7586 & 0.7611 \\
&LangSplat~\cite{qin2024langsplat}& 0.5545 & 0.8071 & 0.8600 \\
&OpenNeRF~\cite{engelmann2024opennerf} & 0.5734 & 0.8345 & 0.8959 \\
&LSM~\cite{fan2024large}&0.6254&0.8887&0.9192 \\
&GOI~\cite{qu2024goi}& 0.8646 & 0.9569 & 0.9362 \\
&\textbf{PE3R}, \emph{ours}       & \textbf{0.8951} & \textbf{0.9617} & \textbf{0.9726} \\
\midrule
\multirow{5}{*}{Rep.}
&LERF~\cite{LERF}                 & 0.2815 & 0.7071 & 0.6602 \\
&F-3DGS~\cite{feature3dgs}  & 0.4480 & 0.7901 & 0.7310 \\
&GS Grouping~\cite{gaussianGroup} & 0.4170 & 0.7370 & 0.7276 \\
&LangSplat~\cite{qin2024langsplat}& 0.4703 & 0.7694 & 0.7604 \\
&OpenNeRF~\cite{engelmann2024opennerf} & 0.5014 & 0.7915 & 0.7831 \\
&LSM~\cite{fan2024large}&0.5614&0.8113&0.7961 \\
&GOI~\cite{qu2024goi}& 0.6169 & 0.8367 & 0.8088 \\
&\textbf{PE3R}, \emph{ours}       & \textbf{0.6531} & \textbf{0.8377} & \textbf{0.8444} \\
\bottomrule
\end{tabular} \vspace{-5px}
\caption{\textbf{2D-to-3D Open-Vocabulary Segmentation} results on small-scale Mip-NeRF360 (Mip.) and Replica (Rep.) datasets.}
\label{tab1}
\end{table}

\subsection{Semantic Point Cloud Reconstruction}
Building upon the disambiguated pixel embeddings, this module focuses on reconstructing a high-quality 3D semantic point cloud.
We leverage recent feed-forward 3D pointmap predictors, such as DUSt3R~\cite{wang2024dust3r}, to directly estimate per-pixel 3D coordinates (or pointmaps) from the multi-view images:
\begin{equation}
\begin{split}
\label{eq12}
\mathbf{P}^1,\mathbf{P}^2,\ldots,\mathbf{P}^n=\mathcal{F}_{pts}(\mathbf{X}^{1},\mathbf{X}^{2},...,\mathbf{X}^{n}),
\end{split}
\end{equation}
where each $\mathbf{P}^i \in \mathbb{R}^{H \times W \times 3}$ represents the 3D coordinates $(x, y, z)$ for every pixel in the $i$-th view.
While this feed-forward paradigm is fast, the initial pointmaps are often contaminated by spatial noise due to visual artifacts like occlusions, reflections, and transparent surfaces.
To overcome this inherent limitation, we introduce a semantic-guided refinement process that utilizes the consistent embeddings to enforce both geometric and semantic coherence.
\noindent\textbf{Anomaly Point Detection.}
The core premise of our detection strategy is that pixels belonging to the semantic region should form a spatially coherent cluster in 3D.
We thus identify outliers by evaluating the local 3D consistency within semantically homogeneous areas.
For each pixel $P_{i,j}$, we compute its average Euclidean distance to neighboring pixels within a $k \times k$ window that share the same semantic label:
\begin{equation}
\label{eq14}
L_{i,j} = \frac{\sum_{dx,dy} \mathcal{I}(\mathbf{M}_{i,j}, \mathbf{M}_{i+dx,j+dy}) \cdot \mathcal{D}(P_{i,j}, P_{i+dx,j+dy})}{\sum_{dx,dy} \mathcal{I}(\mathbf{M}_{i,j}, \mathbf{M}_{i+dx,j+dy})},
\end{equation}
where $dx, dy \in [-\lfloor k/2 \rfloor, \lfloor k/2 \rfloor]$, $\mathcal{I}(\cdot,\cdot)$ is an indicator function that evaluates to 1 if two pixels share the same instance-track index, and $\mathcal{D}(\cdot,\cdot)$ denotes the Euclidean distance in 3D space.
This operation yields a semantic-aware distance map $\mathbf{L}^i \in \mathbb{R}^{H \times W}$ for each view.
Pixels associated with distance values exceeding a predefined threshold after normalization are flagged as spatial anomalies and subsequently excluded, resulting in a significantly cleaner intermediate point cloud.

\begin{table}[t]
\centering
\tablestyle{7.5pt}{0.8}
\begin{tabular}{c|ccc}
\toprule
 Method & Preprocess & Training & Total \\
\midrule
LERF~\cite{LERF} & 3mins & 40mins & 43mins \\
F-3DGS~\cite{feature3dgs}  & 25mins & 623mins & 648mins  \\
GS Grouping~\cite{gaussianGroup} & 27mins & 138mins & 165mins \\
LangSplat~\cite{qin2024langsplat}& 50mins & 99mins & 149mins \\
GOI~\cite{qu2024goi}& 8mins & 37mins & 45mins \\
\textbf{PE3R}, \emph{ours}       & \textbf{5mins} & \textbf{-} & \textbf{5mins} \\
\bottomrule
\end{tabular} \vspace{-5px}
\caption{\textbf{Runtime Comparison for 3D Semantic Reconstruction} on Mipnerf360.} \label{tab2}
\end{table}

\begin{table}[t]
\centering
\tablestyle{7.0pt}{0.8}
\begin{tabular}{c|ccc}
\toprule
Method & mIoU & mPA & mP \\
\midrule
LERF~\cite{LERF} Embedding    & 0.1824 & 0.6024 & 0.5873 \\
GOI~\cite{qu2024goi} Embedding & 0.2101 & 0.6216 & 0.6013 \\
LSM~\cite{fan2024large} Embedding & 0.2124 & 0.6346 & 0.6114 \\
\textbf{PE3R Embedding (ours)}          & \textbf{0.2248} & \textbf{0.6542} & \textbf{0.6315} \\
\bottomrule
\end{tabular} \vspace{-5px}
\caption{\textbf{2D-to-3D Open-Vocabulary Segmentation} results on the large-scale ScanNet++ dataset.} \label{tab3}
\end{table}

\noindent\textbf{Semantic-Guided Refinement.}
Merely filtering outliers, however, does not actively correct the underlying erroneous estimates.
Instead of applying computationally expensive post-hoc geometric regularization (\eg, least-squares fitting), we propose an efficient image-space smoothing strategy that tackles the problem at its root.
For each detected anomaly, we adjust its RGB value $\mathbf{y}$ by blending it with the average color $\mathbf{x}$ of its surrounding semantic region:
\begin{equation}
\hat{y} = \alpha \cdot x + (1-\alpha) \cdot y.
\end{equation}
Here, the blending factor $\alpha \in [0, 1]$ controls the strength of semantic smoothing, effectively suppressing textural noise that misleads the geometric predictor while preserving genuine structural edges.
The smoothed image is then fed back into the pointmap predictor $\mathcal{F}_{pts}$, yielding a refined 3D estimate that is more spatially coherent and better aligned with the semantic structure of the scene.
Finally, a global alignment step synchronizes all refined pointmaps ${\mathbf{\hat{P}}^i}$ into a unified coordinate frame, ensuring multi-view consistency. The output is a fused semantic point cloud where each point is associated with a 3D coordinate and its corresponding semantic embedding, forming a robust foundation for language-grounded interaction.

\begin{table*}[t]
\tablestyle{7.5pt}{0.8}
\centering
\begin{tabular}{x{7}l|cc|cc|cc|cc|cc|cc}
\toprule
\multicolumn{2}{c|}{\multirow{2}{*}{Methods}} & \multicolumn{2}{c|}{KITTI} & \multicolumn{2}{c|}{ScanNet} & \multicolumn{2}{c|}{ETH3D} & \multicolumn{2}{c|}{DTU} & \multicolumn{2}{c|}{T\&T} & \multicolumn{2}{c}{Ave.} \\ \cline{3-14}
\multicolumn{2}{c|}{} & rel$\downarrow$ & $\tau\uparrow$ & rel$\downarrow$ & $\tau\uparrow$ & rel$\downarrow$ & $\tau\uparrow$ & rel$\downarrow$ & $\tau\uparrow$ & rel$\downarrow$ & $\tau\uparrow$ & rel$\downarrow$ & $\tau\uparrow$ \\
\midrule
\multirow{2}{*}{(a)} & COLMAP~\cite{schonberger2016structure} &{12.0}&{58.2} &{14.6}&{34.2}&{16.4}&{55.1}&{0.7}&{96.5}&{2.7}& {95.0} &{9.3} & {67.8} \\
& COLMAP Dense~\cite{schonberger2016pixelwise} & 26.9 & 52.7 & 38.0 & 22.5 & 89.8 & 23.2 & 20.8 & 69.3 & 25.7 & 76.4 & 40.2 & 48.8 \\
\midrule
\multirow{4}{*}{(b)} & DUSt3R~\cite{wang2024dust3r} & {9.1} &39.5& 4.9 & 60.2 &2.9&76.9& 3.5&69.3 &3.2&76.7&4.7&64.5 \\
&Fast3R~\cite{Yang_2025_Fast3R} & - & - & 6.3 & 50.3 & 4.7 & 62.7 & 3.9 & 62.6 & 4.4 & 64.0 & - & -  \\
& DUSt3R~\cite{wang2024dust3r}$^\dagger$ & 11.0 & 33.2 & 4.8 & { 60.3} & 3.1 & 74.5 & {2.7} & {75.7} & 2.9 & 78.5 & 4.9 & 64.4 \\
& MASt3R~\cite{mast3r}$^\dagger$& 36.9 & 5.4 & 22.0 & 9.6 & 27.9 & 9.9 & 13.6 & 13.7 & 22.1 & 14.6 & 24.5 & 10.6 \\
& VGGT~\cite{wang2025vggt}& 9.4 & 41.3 & {4.4} & 65.0 & 1.8 & 86.3 & 1.1 & 94.3 & 2.1 & 85.2 & 3.8 & 74.4 \\
\midrule
\multirow{2}{*}{(c)} & \textbf{PE3R}, with DUSt3R & 9.4 & {48.6} & 5.5 & 55.1 & {2.3} & {82.0} & 3.2 & 69.1 & {2.1} & {85.3} & {4.5} & {68.0} \\
& \textbf{PE3R}, with VGGT & 9.2 & 43.8 & 4.3 & 65.8 & 1.7 & 86.6 & 1.1 & 94.4 & 2.0 & 85.4 & 3.7 & 75.2 \\
\bottomrule
\end{tabular} \vspace{-5px}
\caption{\textbf{Multi-view Depth Estimation on RobustMVD.}
Comparison under: (a) classical methods, (b) feed-forward methods, and (c) semantic reconstruction.
PE3R achieves the best average performance, demonstrating that semantic guidance generally enhances geometric accuracy.
$^\dagger$ indicates our implementation.
Metrics are indicated by relative error (rel \(\downarrow\)) and inlier ratio (\(\tau \uparrow\)).
} \label{tab4}
\end{table*}

\subsection{Global View Perception}

The final stage of PE3R leverages the unified semantic point cloud to enable intuitive, open-vocabulary interaction with the reconstructed 3D semantic scene.
Given a user-provided text query, we first encode it into a semantic embedding using a language encoder $\mathcal{F}_{txt}(\cdot)$ (\eg, CLIP~\cite{radford2021learning}):
\begin{equation}
\label{eq15}
\mathbf{T} = \mathcal{F}_{txt}(\text{text}),
\end{equation}
where $\mathbf{T} \in \mathbb{R}^d$ resides in the shared image-language embedding space.
To identify corresponding regions in 3D, we compute the cosine similarity between the query embedding $\mathbf{T}$ and the dense per-pixel semantic embeddings $\mathbf{E}^1, \mathbf{E}^2, \ldots, \mathbf{E}^n$ from all input views:
\begin{equation}
\label{eq16}
[\mathbf{S}^1, \mathbf{S}^2, \ldots, \mathbf{S}^n] = \mathcal{D}(\mathbf{T}, [\mathbf{E}^1, \mathbf{E}^2, \ldots, \mathbf{E}^n]),
\end{equation}
where $\mathcal{D}(\cdot)$ denotes the cosine similarity function.
To ensure consistency in retrieval scores across different viewpoints, we apply a global min–max normalization to all similarity values.
Specifically, we concatenate the similarity scores from every view and scale them globally into the range $[0, 1]$.
This provides a unified and interpretable similarity measure that is invariant to view-specific variations.
A predefined threshold is then applied to the normalized scores to select the most relevant 3D points from the corresponding refined pointmaps $\mathbf{P}^1, \mathbf{P}^2, \ldots, \mathbf{P}^n$.
Points whose normalized similarity scores exceed this threshold are considered semantically aligned with the text query and are retained for 3D localization.
This approach enables natural language-based interaction with the reconstructed 3D scene, supporting flexible object retrieval and semantic scene understanding without the need for predefined category labels or model retraining.

\section{Experiments}
\label{sec:experiments}

\subsection{Experimental Details}
\noindent\textbf{Datasets and Tasks.}
We evaluate PE3R on two core tasks to comprehensively assess its capabilities: \textit{open-vocabulary 2D-to-3D segmentation} and \textit{multi-view depth estimation}.
For open-vocabulary segmentation, we employ three benchmarks:
Mip-NeRF360~\cite{barron2022mip} and Replica~\cite{straub2019replica} are used to evaluate performance in relatively compact, object-centric scenes, using the open-vocabulary labels from GOI~\cite{qu2024goi}.
ScanNet++~\cite{yeshwanth2023scannet++} is adopted to test generalization in large-scale, complex indoor environments with diverse layouts and object categories.
For multi-view depth estimation, we follow established benchmarks to assess geometric accuracy across a wide range of scenarios.
The evaluated datasets include KITTI~\cite{geiger2013vision} (outdoor driving), ScanNet~\cite{dai2017scannet} (indoor scenes), DTU~\cite{aanaes2016large} (controlled object-level scans), ETH3D~\cite{schops2017multi} (high-resolution scenes), and Tanks and Temples (T\&T)~\cite{knapitsch2017tanks} (large-scale outdoor).
This selection ensures a comprehensive evaluation across both indoor and outdoor environments.

\noindent\textbf{Implementation Details.}
PE3R is built upon several powerful, off-the-shelf models.
We employ MobileSAMv2~\cite{mobile_sam_v2} for efficient image segmentation, SAM2~\cite{sam2} for obtaining temporally consistent mask tracks, and DUSt3R~\cite{wang2024dust3r} as our core feed-forward 3D pointmap predictor.
We use SAM for per-image ``segment-everything'' masks, and SAM2 for instance tracking.
Concretely, we initialize SAM2 on the first image by running SAM and using the resulting masks as prompts to seed the SAM2 tracker.
For each subsequent image, we (i) run SAM2 to propagate tracked masks, then (ii) run SAM to re-detect all objects; any SAM mask with IoU$<$0.1 to \emph{all} current SAM2 masks is treated as newly discovered/re-detected and added as a new instance to the tracker, recovering missed/lost tracks under large viewpoint changes.
All experiments were conducted on a single server equipped with an NVIDIA A100 GPU and an Intel i7 CPU.

\noindent\textbf{Evaluation Protocol.}
Our evaluation protocol is designed to be fair, reproducible, and aligned with standard practices in the field.
For open-vocabulary segmentation, we report the standard metrics of mean Intersection over Union (mIoU), mean Pixel Accuracy (mPA), and mean Precision (mP).
The evaluation procedure is tailored to dataset characteristics to ensure optimal and fair benchmarking.
On Mip-NeRF360 and Replica, we follow the protocol established by GOI~\cite{qu2024goi}.
We perform a holistic 3D semantic reconstruction of the entire scene.
Open-vocabulary queries, labeled by GOI, are then executed in this unified 3D space.
The resulting 3D segmentation is projected back onto the original 2D image planes for quantitative comparison against the 2D ground-truth masks.
On the larger ScanNet++ dataset, we process data in snippets of four consecutive frames to reduce the computational load, a common practice for scaling 3D methods to large scenes.
This approach effectively highlights the benefit of integrating semantic features into 3D reconstruction.
For a direct and fair comparison, we integrate the semantic embeddings from baseline methods, LERF~\cite{LERF}, LSM~\cite{fan2024large}, and GOI~\cite{qu2024goi}, directly into our pipeline.
This is done by replacing \textit{only} the pixel-level semantic embeddings while keeping the geometry backbone and the entire evaluation procedure identical.
The final evaluation is conducted by projecting the 3D segmentation results back to 2D for comparison with ground-truth annotations.
This controlled setup ensures that performance differences are attributable solely to the quality of the semantic embeddings.
We report the total pipeline processing time (preprocessing and inference) to assess computational efficiency.
For multi-view depth estimation, we adhere to the official protocol of the RobustMVD benchmark~\cite{schroeppel2022robust}.
For each test sample, we use source views selected via the quasi-optimal strategy.
Performance is measured using Absolute Relative Error (Rel) $\downarrow$ and the Inlier Ratio ($\tau$) $\uparrow$ at a threshold of $1.03$.
Results are reported for each individual dataset as well as the overall average across all benchmarks.

\vspace{-1mm}
\subsection{Main Results}

\begin{table}[t]
\tablestyle{6.5pt}{0.8}
\centering
\begin{tabular}{c|ccc}
\toprule
Method & mIoU & mPA & mP \\
\midrule
w/o Multi-Level Disam.      & 0.1624 & 0.5892 & 0.5623 \\
w/o Cross-View Disam.       & 0.1895 & 0.6012 & 0.5923 \\
w/o Global MinMax Norm.     & 0.2035 & 0.6253 & 0.6186 \\
\textbf{PE3R (full)}        & \textbf{0.2248} & \textbf{0.6542} & \textbf{0.6315} \\
\bottomrule
\end{tabular} \vspace{-5px}
\caption{\textbf{Effect of Disambiguation Modules} for 2D-to-3D open-vocabulary segmentation on ScanNet++.} \label{tab5}
\vspace{-5px}
\end{table}

\begin{figure}[t]
\begin{center}
\centerline{\includegraphics[width=0.96\linewidth]{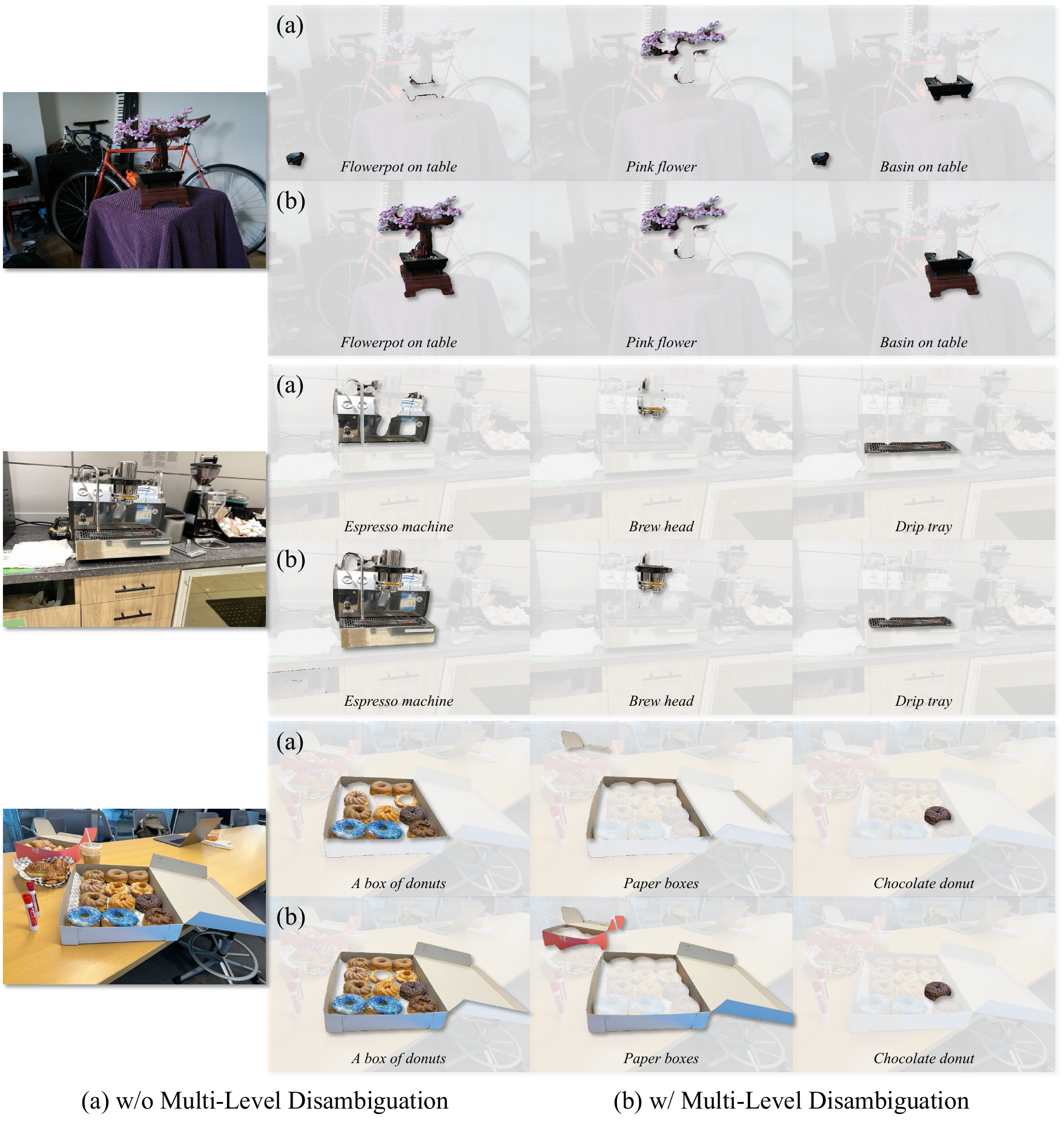}}\vspace{-10px}
\caption{\textbf{Effect of Multi-Level Disambiguation.}
Without this module, the model fails to associate object parts with the whole, leading to fragmented semantics.
}
\label{fig3}
\end{center}
\vspace{-20px}
\end{figure}

\noindent\textbf{2D-to-3D Open-Vocabulary Segmentation.}
We first evaluate PE3R on the Mip-NeRF360 and Replica datasets, which provide a standard benchmark for comparing with state-of-the-art NeRF- and 3DGS-based methods.
As summarized in Table~\ref{tab1}, PE3R consistently surpasses all previous approaches, establishing a new state-of-the-art across all metrics (mIoU, mPA, mP).
A pivotal advantage of our framework is its exceptional computational efficiency.
As evidenced by Table~\ref{tab2}, PE3R completes the full semantic reconstruction pipeline in merely 5 minutes.
This represents up to a 9$\times$ speedup over optimization-based methods.
To assess generalization, we evaluate PE3R on the large-scale and complex ScanNet++ benchmark.
Given the scalability constraints of methods requiring per-scene optimization, we conduct a direct and fair comparison by integrating the 2D semantic features from LERF, LSM, and GOI into our pipeline under identical settings.
As reported in Table~\ref{tab3}, PE3R achieves superior segmentation accuracy, confirming its ability to maintain high performance while scaling effectively to challenging, large-scale environments.
Qualitative results in Fig.~\ref{fig1} further corroborate these findings, showcasing robust and precise segmentation across diverse indoor, outdoor, and in-the-wild scenes.

\begin{figure}[t]
\begin{center}
\centerline{\includegraphics[width=0.96\linewidth]{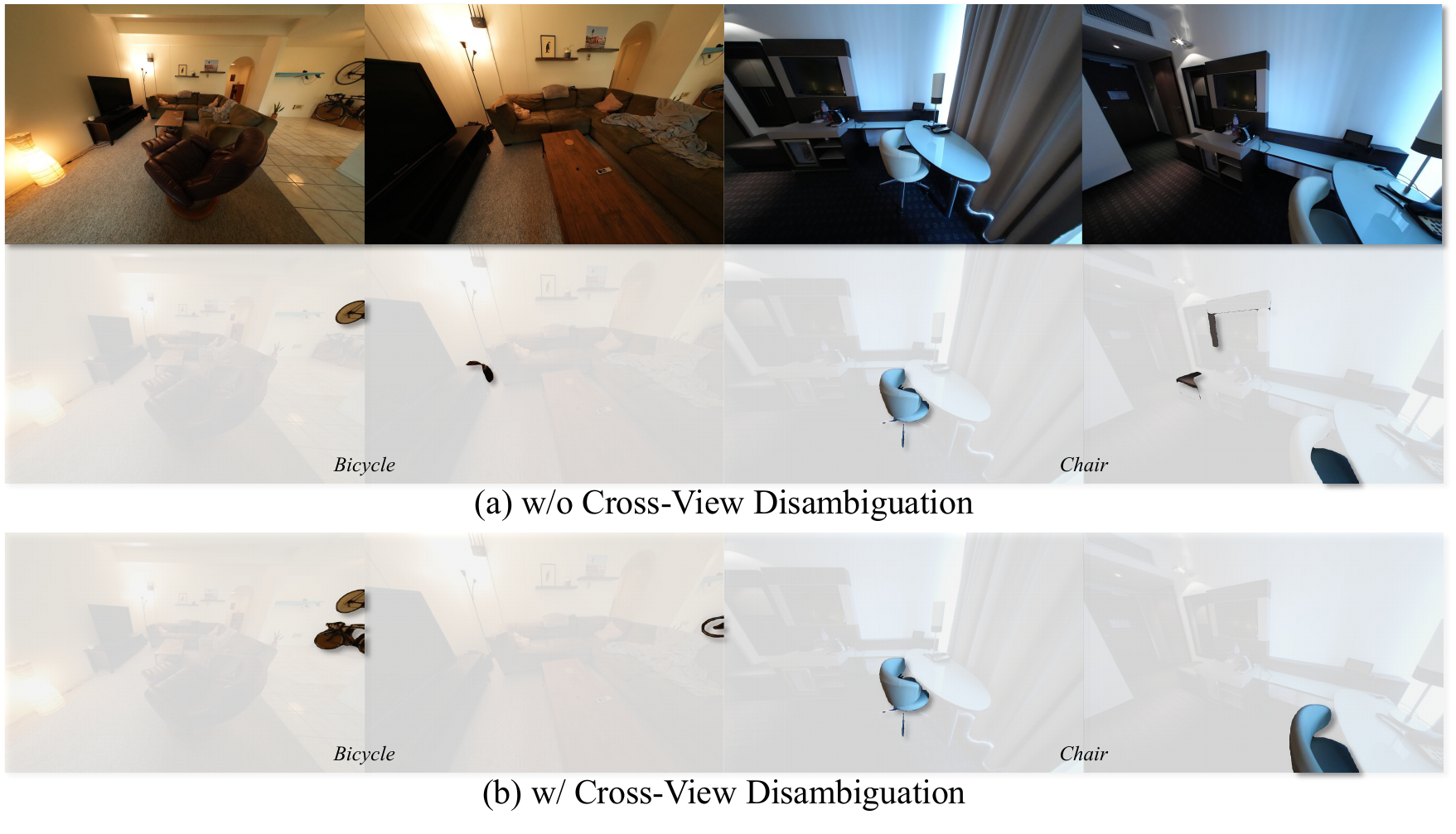}} \vspace{-10px}
\caption{\textbf{Effect of Cross-View Disambiguation.}
In the absence of cross-view alignment, the same object is inconsistently labeled across viewpoints.
}
\label{fig4}
\end{center}
\vspace{-35px}
\end{figure}

\begin{figure}[t]
\begin{center}
\centerline{\includegraphics[width=1\linewidth]{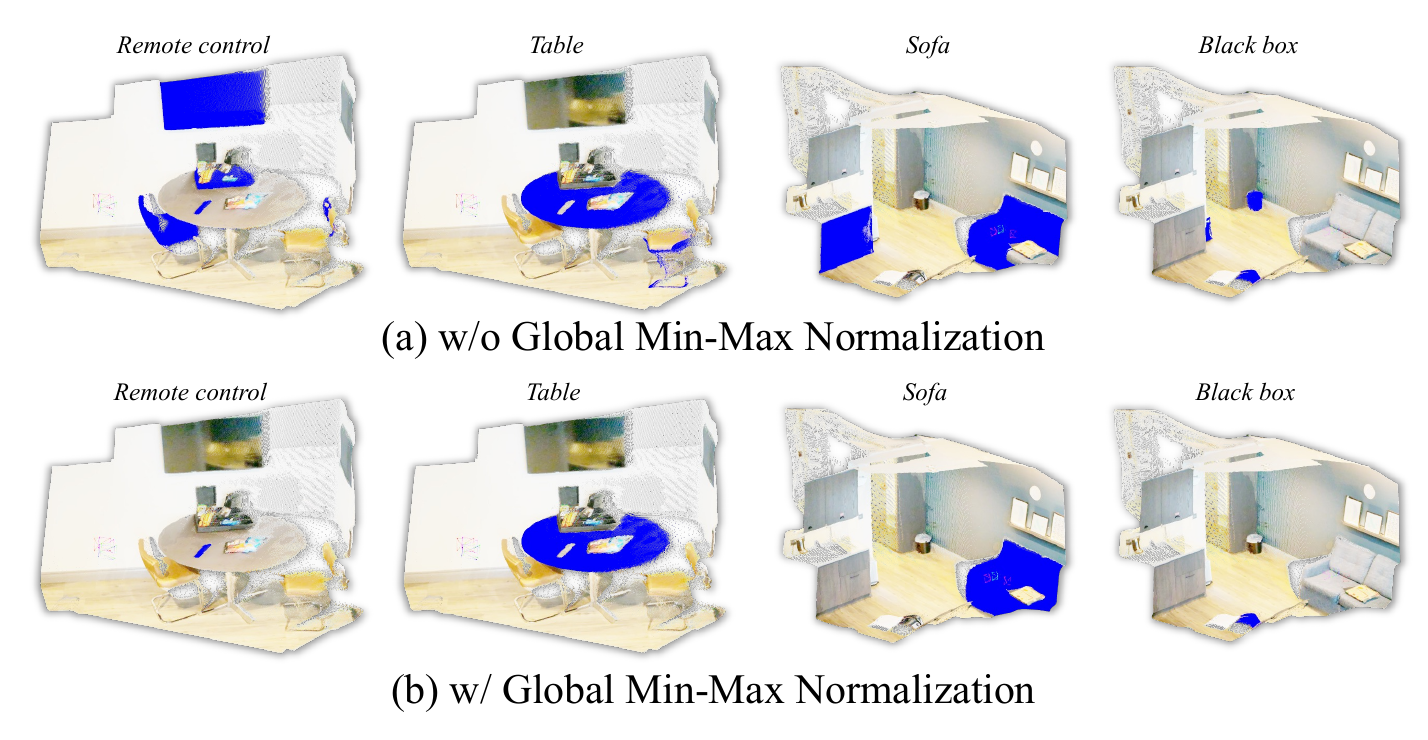}} \vspace{-10px}
\caption{\textbf{Effect of Global Min-Max Normalization.}
Without global normalization, similarity scores become uncalibrated across views, impairing retrieval reliability.
}
\label{fig5}
\end{center}
\vspace{-20px}
\end{figure}
\noindent\textbf{Multi-view Depth Estimation.}
We further evaluate the geometric fidelity of PE3R on the task of multi-view depth estimation.
For a comprehensive analysis, Table~\ref{tab4} organizes methods into three categories: (a) classical 3D reconstruction approaches, (b) modern feed-forward methods, and (c) our full PE3R framework with semantic-guided reconstruction.
PE3R demonstrates highly competitive performance across a wide spectrum of datasets, achieving the best overall average on standard metrics (Rel and $\tau$).
This indicates that the integration of semantic priors generally enhances the robustness of the geometric reconstruction.
An in-depth analysis reveals that the benefits of semantic guidance are scene-dependent.
While PE3R shows notable improvements on most benchmarks, its performance on ScanNet and DTU is marginally behind the DUSt3R baseline.
We attribute this to the characteristic challenges of these datasets, such as low semantic diversity, repetitive textures, and limited viewpoint coverage, where semantic cues can introduce ambiguity rather than resolve it.
This nuanced performance profile highlights a key insight: semantic guidance is most beneficial in semantically rich and visually distinct environments.
Nevertheless, the improvement in the overall benchmark average strongly validates that our semantic augmentation strategy provides a net positive effect on reconstruction quality.

\begin{table}[t]
\tablestyle{2.0pt}{0.8}
\centering
\begin{tabular}{c|c|ccc}
\toprule
Multi-Level Disam. & Cross-View Disam. & mIoU & mPA & mP \\
\midrule
averaged & averaged   & 0.1994 & 0.6001 & 0.5985 \\
averaged & area-slerp & 0.2158 & 0.6141 & 0.6015 \\
area-slerp & averaged & 0.2058 & 0.6041 & 0.6003 \\ 
conf-weighted & conf-weighted & 0.1532 & 0.5723 & 0.5435 \\ 
conf-weighted & conf-slerp & 0.1743 & 0.5842 & 0.5554 \\ 
conf-slerp & conf-weighted & 0.1721 & 0.5813 & 0.5493 \\ 
conf-slerp & conf-slerp & 0.1998 & 0.5991 & 0.5918 \\ \midrule
area-slerp & area-slerp  & \textbf{0.2248} & \textbf{0.6542} & \textbf{0.6315} \\
\bottomrule
\end{tabular} \vspace{-5px}
\caption{\textbf{Comparison of Feature Aggregation Strategies} on ScanNet++.
%
} \label{tab6}
\end{table}

\subsection{Ablation Studies}
To provide a comprehensive analysis of PE3R's design, we present key ablation studies here, with additional experiments detailed in the supplementary material.
All evaluations are conducted on both open-vocabulary segmentation and multi-view depth estimation to quantitatively assess the contribution of each component.

\noindent\textbf{Effect of Disambiguation Modules.}
We first analyze the three core modules responsible for establishing a consistent semantic space: Multi-Level Disambiguation, Cross-View Disambiguation, and Global Min-Max Normalization.
Quantitative results on ScanNet++ (Table~\ref{tab5}) demonstrate that ablating any module causes a performance drop, with the most significant degradation occurring when Multi-Level Disambiguation is removed.
This underscores its critical role in resolving hierarchical ambiguities and maintaining part-whole consistency.
Qualitative results in Figs.~\ref{fig3}--\ref{fig5} offer further insight.
As shown in Fig.~\ref{fig3}, the absence of Multi-Level Disambiguation leads to fragmented semantics.
While parts (\eg, ``Drip tray'') may be detected, the model fails to associate them with the whole object (\eg, ``Espresso machine'').
Our module enables coherent hierarchical understanding.
Fig.~\ref{fig4} illustrates that without Cross-View Disambiguation, the same object (\eg, ``Bicycle'' or ``Chair'') is inconsistently labeled across viewpoints due to occlusion or perspective changes.
Integrating this module restores semantic coherence throughout the 3D scene.
Fig.~\ref{fig5} reveals that removing Global Min-Max Normalization results in uncalibrated similarity scores across views, impairing retrieval reliability.
This step ensures a consistent and interpretable similarity range, which is crucial for robust open-vocabulary localization.
To validate our specific design choice for feature aggregation, we compare our geometry-aware interpolation against naive uniform averaging and confidence-weighted baselines using DUSt3R's per-point confidence (conf-avg and conf-slerp).
As shown in Table~\ref{tab6}, uniform averaging leads to consistently lower mIoU and less accurate retrieval, while our area-weighted slerp outperforms both confidence-weighted variants on ScanNet++.
This confirms that our area-weighted, spherical interpolation is essential for generating discriminative and reliable semantic embeddings.

\begin{table}[t]
\tablestyle{5.5pt}{0.8}
\centering
\begin{tabular}{c|cccc}
\toprule
Method & rel$\downarrow$ & $\tau\uparrow$ & Runtime & $\Delta$ \\
\midrule
w/o Semantic Recon. & 4.9 & 64.4 & 10.40s & - \\
\textbf{PE3R (full)} & \textbf{4.5} & \textbf{68.0} & 11.19s & +0.79s \\
\bottomrule
\end{tabular} \vspace{-5px}
\caption{\textbf{Ablation on Semantic Point Cloud Reconstruction.}
Removing this module degrades geometric quality.} \label{tab7}
\vspace{-10px}
\end{table}

\begin{figure}[t]
\begin{center}
\centerline{\includegraphics[width=1\linewidth]{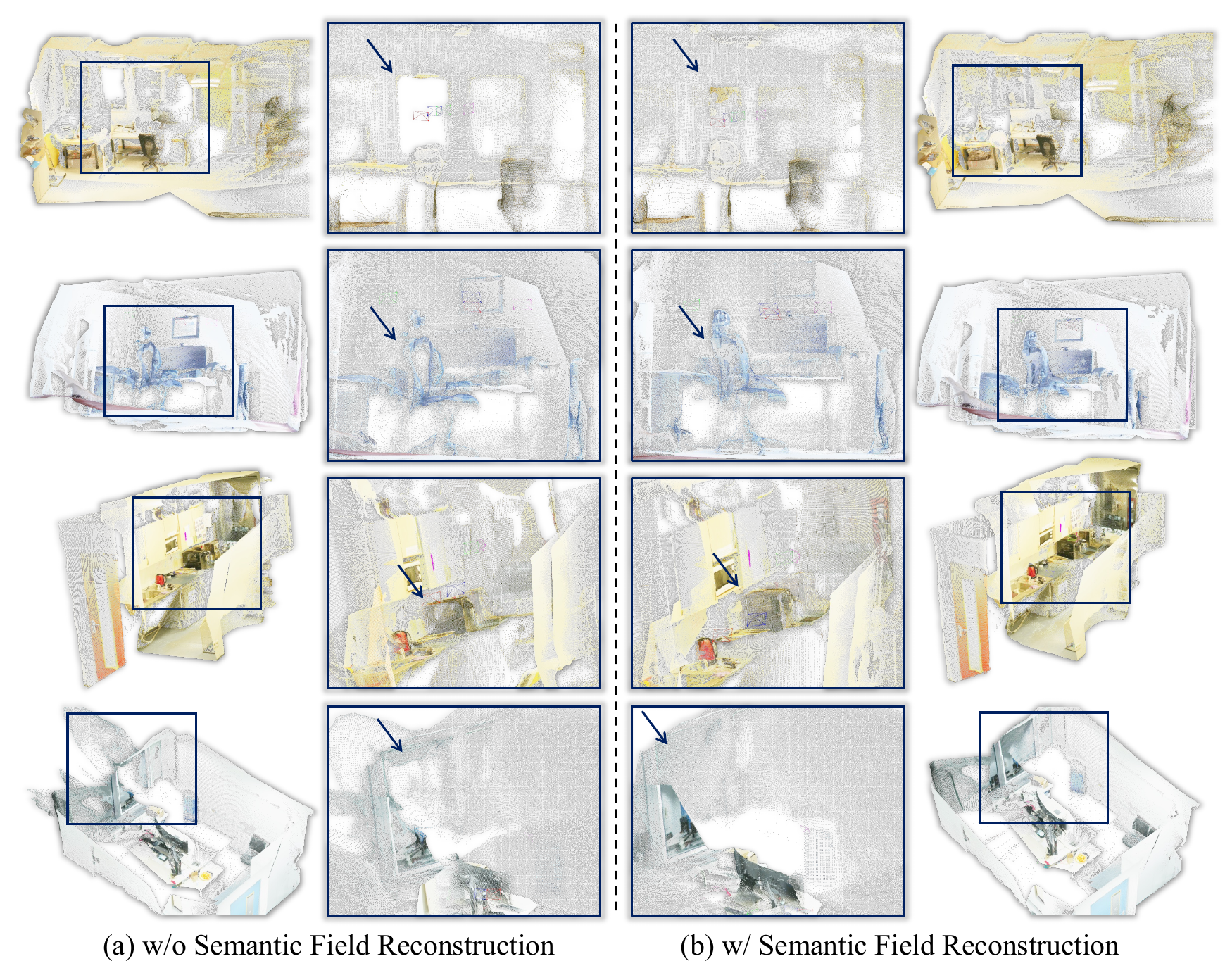}}\vspace{-10px}
\caption{\textbf{Effect of Semantic Point Cloud Reconstruction.}
Removing this module increases spatial noise and reduces object-level coherence in the reconstructed point cloud, highlighting its importance for both geometric and semantic fidelity.}
\label{fig6}
\end{center}
\vspace{-20px}
\end{figure}

\noindent\textbf{Effect of Semantic Reconstruction Module.}
We further ablate the Semantic Point Cloud Reconstruction module to evaluate its impact on geometric accuracy.
The results in Table~\ref{tab7} show that this module brings a noticeable improvement in depth estimation metrics with only a minimal computational overhead.
Qualitatively, Fig.~\ref{fig6} demonstrates that semantic-guided refinement effectively suppresses spatial outliers caused by challenging conditions like reflections and transparency, resulting in cleaner and more structurally coherent point clouds.
This confirms that semantic cues actively contribute to enhancing the underlying geometry, beyond merely providing labels.
\section{Conclusion}
\label{sec:conclusion}

We presented PE3R, a framework that establishes a new paradigm for efficient and generalizable 3D semantic reconstruction from unposed images.
By integrating pixel embedding disambiguation, semantic point cloud reconstruction, and global view perception into a cohesive pipeline, PE3R achieves robust zero-shot generalization across diverse scenes without any scene-specific fine-tuning.
Extensive evaluations across diverse benchmarks demonstrate that PE3R achieves state-of-the-art accuracy with good efficiency.
We believe PE3R paves the way for more scalable and practical 3D vision systems, with promising applications in robotics, AR/VR, and autonomous navigation.

\paragraph{Acknowledgements.} 
This project is supported by the National Research Foundation, Singapore, under its Medium Sized Center for Advanced Robotics Technology Innovation.

{
    \small
    \bibliographystyle{ieeenat_fullname}
    \bibliography{main}

@String(CVPR= {IEEE Conf. Comput. Vis. Pattern Recog.})

@String(TOG= {ACM Trans. Graph.})

@String(CVPR  = {CVPR})

@String(TOG   = {ACM TOG})

@article{ayzenberg2024development,
  title={Development of visual object recognition},
  author={Ayzenberg, Vladislav and Behrmann, Marlene},
  journal={Nature Reviews Psychology},
  volume={3},
  number={2},
  pages={73--90},
  year={2024},
  publisher={Nature Publishing Group US New York}
}

@article{sinha1996role,
  title={Role of learning in three-dimensional form perception},
  author={Sinha, Pawan and Poggio, Tomaso},
  journal={Nature},
  volume={384},
  number={6608},
  pages={460--463},
  year={1996},
  publisher={Nature Publishing Group UK London}
}

@article{welchman20053d,
  title={3D shape perception from combined depth cues in human visual cortex},
  author={Welchman, Andrew E and Deubelius, Arne and Conrad, Verena and B{\"u}lthoff, Heinrich H and Kourtzi, Zoe},
  journal={Nature neuroscience},
  volume={8},
  number={6},
  pages={820--827},
  year={2005},
  publisher={Nature Publishing Group US New York}
}

@article{sa3d,
  title={Segment anything in 3d with nerfs},
  author={Cen, Jiazhong and Zhou, Zanwei and Fang, Jiemin and Shen, Wei and Xie, Lingxi and Jiang, Dongsheng and Zhang, Xiaopeng and Tian, Qi and others},
  journal={Advances in Neural Information Processing Systems},
  volume={36},
  pages={25971--25990},
  year={2023}
}

@article{nerf,
  title={Nerf: Representing scenes as neural radiance fields for view synthesis},
  author={Mildenhall, Ben and Srinivasan, Pratul P and Tancik, Matthew and Barron, Jonathan T and Ramamoorthi, Ravi and Ng, Ren},
  journal={Communications of the ACM},
  volume={65},
  number={1},
  pages={99--106},
  year={2021},
  publisher={ACM New York, NY, USA}
}

@article{3dgs,
  title={3D Gaussian Splatting for Real-Time Radiance Field Rendering.},
  author={Kerbl, Bernhard and Kopanas, Georgios and Leimk{\"u}hler, Thomas and Drettakis, George},
  journal={ACM Trans. Graph.},
  volume={42},
  number={4},
  pages={139--1},
  year={2023}
}

@inproceedings{liu2024sanerf,
  title={SANeRF-HQ: Segment Anything for NeRF in High Quality},
  author={Liu, Yichen and Hu, Benran and Tang, Chi-Keung and Tai, Yu-Wing},
  booktitle={Proceedings of the IEEE/CVF Conference on Computer Vision and Pattern Recognition},
  pages={3216--3226},
  year={2024}
}

@article{semanticAnyIn3dgs,
  title={Semantic anything in 3d gaussians},
  author={Hu, Xu and Wang, Yuxi and Fan, Lue and Fan, Junsong and Peng, Junran and Lei, Zhen and Li, Qing and Zhang, Zhaoxiang},
  journal={arXiv preprint arXiv:2401.17857},
  year={2024}
}

@article{gaussianGroup,
  title={Gaussian grouping: Segment and edit anything in 3d scenes},
  author={Ye, Mingqiao and Danelljan, Martin and Yu, Fisher and Ke, Lei},
  journal={arXiv preprint arXiv:2312.00732},
  year={2023}
}

@inproceedings{sam,
  title={Segment anything},
  author={Kirillov, Alexander and Mintun, Eric and Ravi, Nikhila and Mao, Hanzi and Rolland, Chloe and Gustafson, Laura and Xiao, Tete and Whitehead, Spencer and Berg, Alexander C and Lo, Wan-Yen and others},
  booktitle={Proceedings of the IEEE/CVF International Conference on Computer Vision},
  pages={4015--4026},
  year={2023}
}

@inproceedings{feature3dgs,
  title={Feature 3dgs: Supercharging 3d gaussian splatting to enable distilled feature fields},
  author={Zhou, Shijie and Chang, Haoran and Jiang, Sicheng and Fan, Zhiwen and Zhu, Zehao and Xu, Dejia and Chari, Pradyumna and You, Suya and Wang, Zhangyang and Kadambi, Achuta},
  booktitle={Proceedings of the IEEE/CVF Conference on Computer Vision and Pattern Recognition},
  pages={21676--21685},
  year={2024}
}

@inproceedings{wang2024dust3r,
  title={Dust3r: Geometric 3d vision made easy},
  author={Wang, Shuzhe and Leroy, Vincent and Cabon, Yohann and Chidlovskii, Boris and Revaud, Jerome},
  booktitle={Proceedings of the IEEE/CVF Conference on Computer Vision and Pattern Recognition},
  pages={20697--20709},
  year={2024}
}

@article{sam2,
  title={Sam 2: Segment anything in images and videos},
  author={Ravi, Nikhila and Gabeur, Valentin and Hu, Yuan-Ting and Hu, Ronghang and Ryali, Chaitanya and Ma, Tengyu and Khedr, Haitham and R{\"a}dle, Roman and Rolland, Chloe and Gustafson, Laura and others},
  journal={arXiv preprint arXiv:2408.00714},
  year={2024}
}

@misc{SAGA,
      title={Segment Any 3D Gaussians}, 
      author={Jiazhong Cen and Jiemin Fang and Chen Yang and Lingxi Xie and Xiaopeng Zhang and Wei Shen and Qi Tian},
      year={2024},
      eprint={2312.00860},
      archivePrefix={arXiv},
      primaryClass={cs.CV},
      url={https://arxiv.org/abs/2312.00860}, 
}

@article{chen2023gnesf,
  title={Gnesf: Generalizable neural semantic fields},
  author={Chen, Hanlin and Li, Chen and Guo, Mengqi and Yan, Zhiwen and Lee, Gim Hee},
  journal={Advances in Neural Information Processing Systems},
  volume={36},
  pages={36553--36565},
  year={2023}
}

@inproceedings{inplace,
  title={In-place scene labelling and understanding with implicit scene representation},
  author={Zhi, Shuaifeng and Laidlow, Tristan and Leutenegger, Stefan and Davison, Andrew J},
  booktitle={Proceedings of the IEEE/CVF International Conference on Computer Vision},
  pages={15838--15847},
  year={2021}
}

@inproceedings{liu2023instance,
  title={Instance neural radiance field},
  author={Liu, Yichen and Hu, Benran and Huang, Junkai and Tai, Yu-Wing and Tang, Chi-Keung},
  booktitle={Proceedings of the IEEE/CVF International Conference on Computer Vision},
  pages={787--796},
  year={2023}
}

@article{contrastiveLift,
  title={Contrastive lift: 3d object instance segmentation by slow-fast contrastive fusion},
  author={Bhalgat, Yash and Laina, Iro and Henriques, Joao F and Zisserman, Andrew and Vedaldi, Andrea},
  journal={arXiv preprint arXiv:2306.04633},
  year={2023}
}

@inproceedings{panopticLift,
  title={Panoptic lifting for 3d scene understanding with neural fields},
  author={Siddiqui, Yawar and Porzi, Lorenzo and Bul{\'o}, Samuel Rota and M{\"u}ller, Norman and Nie{\ss}ner, Matthias and Dai, Angela and Kontschieder, Peter},
  booktitle={Proceedings of the IEEE/CVF Conference on Computer Vision and Pattern Recognition},
  pages={9043--9052},
  year={2023}
}

@inproceedings{giraffe,
  title={Giraffe: Representing scenes as compositional generative neural feature fields},
  author={Niemeyer, Michael and Geiger, Andreas},
  booktitle={Proceedings of the IEEE/CVF Conference on Computer Vision and Pattern Recognition},
  pages={11453--11464},
  year={2021}
}

@article{yu2021unsupervised,
  title={Unsupervised discovery of object radiance fields},
  author={Yu, Hong-Xing and Guibas, Leonidas J and Wu, Jiajun},
  journal={arXiv preprint arXiv:2107.07905},
  year={2021}
}

@inproceedings{ISRF,
  title={Interactive segmentation of radiance fields},
  author={Goel, Rahul and Sirikonda, Dhawal and Saini, Saurabh and Narayanan, PJ},
  booktitle={Proceedings of the IEEE/CVF Conference on Computer Vision and Pattern Recognition},
  pages={4201--4211},
  year={2023}
}

@inproceedings{LERF,
  title={Lerf: Language embedded radiance fields},
  author={Kerr, Justin and Kim, Chung Min and Goldberg, Ken and Kanazawa, Angjoo and Tancik, Matthew},
  booktitle={Proceedings of the IEEE/CVF International Conference on Computer Vision},
  pages={19729--19739},
  year={2023}
}

@article{kobayashi2022decomposing,
  title={Decomposing nerf for editing via feature field distillation},
  author={Kobayashi, Sosuke and Matsumoto, Eiichi and Sitzmann, Vincent},
  journal={Advances in Neural Information Processing Systems},
  volume={35},
  pages={23311--23330},
  year={2022}
}

@article{choi2024click,
  title={Click-Gaussian: Interactive Segmentation to Any 3D Gaussians},
  author={Choi, Seokhun and Song, Hyeonseop and Kim, Jaechul and Kim, Taehyeong and Do, Hoseok},
  journal={arXiv preprint arXiv:2407.11793},
  year={2024}
}

@article{shen2024flashsplat,
  title={FlashSplat: 2D to 3D Gaussian Splatting Segmentation Solved Optimally},
  author={Shen, Qiuhong and Yang, Xingyi and Wang, Xinchao},
  journal={arXiv preprint arXiv:2409.08270},
  year={2024}
}

@inproceedings{SGISRF,
  title={Scene-generalizable interactive segmentation of radiance fields},
  author={Tang, Songlin and Pei, Wenjie and Tao, Xin and Jia, Tanghui and Lu, Guangming and Tai, Yu-Wing},
  booktitle={Proceedings of the 31st ACM International Conference on Multimedia},
  pages={6744--6755},
  year={2023}
}

@inproceedings{qin2024langsplat,
  title={Langsplat: 3d language gaussian splatting},
  author={Qin, Minghan and Li, Wanhua and Zhou, Jiawei and Wang, Haoqian and Pfister, Hanspeter},
  booktitle={Proceedings of the IEEE/CVF Conference on Computer Vision and Pattern Recognition},
  pages={20051--20060},
  year={2024}
}

@article{knapitsch2017tanks,
  title={Tanks and temples: Benchmarking large-scale scene reconstruction},
  author={Knapitsch, Arno and Park, Jaesik and Zhou, Qian-Yi and Koltun, Vladlen},
  journal={ACM Transactions on Graphics (ToG)},
  volume={36},
  number={4},
  pages={1--13},
  year={2017},
  publisher={ACM New York, NY, USA}
}

@inproceedings{barron2022mip,
  title={Mip-nerf 360: Unbounded anti-aliased neural radiance fields},
  author={Barron, Jonathan T and Mildenhall, Ben and Verbin, Dor and Srinivasan, Pratul P and Hedman, Peter},
  booktitle={Proceedings of the IEEE/CVF conference on computer vision and pattern recognition},
  pages={5470--5479},
  year={2022}
}

@article{mast3r,
  title={Grounding Image Matching in 3D with MASt3R},
  author={Leroy, Vincent and Cabon, Yohann and Revaud, J{\'e}r{\^o}me},
  journal={arXiv preprint arXiv:2406.09756},
  year={2024}
}

@inproceedings{tschernezki2022neural,
  title={Neural feature fusion fields: 3d distillation of self-supervised 2d image representations},
  author={Tschernezki, Vadim and Laina, Iro and Larlus, Diane and Vedaldi, Andrea},
  booktitle={2022 International Conference on 3D Vision (3DV)},
  pages={443--453},
  year={2022},
  organization={IEEE}
}

@inproceedings{peng2023openscene,
  title={Openscene: 3d scene understanding with open vocabularies},
  author={Peng, Songyou and Genova, Kyle and Jiang, Chiyu and Tagliasacchi, Andrea and Pollefeys, Marc and Funkhouser, Thomas and others},
  booktitle={Proceedings of the IEEE/CVF conference on computer vision and pattern recognition},
  pages={815--824},
  year={2023}
}

@article{takmaz2023openmask3d,
  title={Openmask3d: Open-vocabulary 3d instance segmentation},
  author={Takmaz, Ay{\c{c}}a and Fedele, Elisabetta and Sumner, Robert W and Pollefeys, Marc and Tombari, Federico and Engelmann, Francis},
  journal={arXiv preprint arXiv:2306.13631},
  year={2023}
}

@article{fu2022geo,
  title={Geo-neus: Geometry-consistent neural implicit surfaces learning for multi-view reconstruction},
  author={Fu, Qiancheng and Xu, Qingshan and Ong, Yew Soon and Tao, Wenbing},
  journal={Advances in Neural Information Processing Systems},
  volume={35},
  pages={3403--3416},
  year={2022}
}

@article{guo2023streetsurf,
  title={Streetsurf: Extending multi-view implicit surface reconstruction to street views},
  author={Guo, Jianfei and Deng, Nianchen and Li, Xinyang and Bai, Yeqi and Shi, Botian and Wang, Chiyu and Ding, Chenjing and Wang, Dongliang and Li, Yikang},
  journal={arXiv preprint arXiv:2306.04988},
  year={2023}
}

@inproceedings{long2022sparseneus,
  title={Sparseneus: Fast generalizable neural surface reconstruction from sparse views},
  author={Long, Xiaoxiao and Lin, Cheng and Wang, Peng and Komura, Taku and Wang, Wenping},
  booktitle={European Conference on Computer Vision},
  pages={210--227},
  year={2022},
  organization={Springer}
}

@article{wang2021neus,
  title={Neus: Learning neural implicit surfaces by volume rendering for multi-view reconstruction},
  author={Wang, Peng and Liu, Lingjie and Liu, Yuan and Theobalt, Christian and Komura, Taku and Wang, Wenping},
  journal={arXiv preprint arXiv:2106.10689},
  year={2021}
}

@article{wang2022hf,
  title={Hf-neus: Improved surface reconstruction using high-frequency details},
  author={Wang, Yiqun and Skorokhodov, Ivan and Wonka, Peter},
  journal={Advances in Neural Information Processing Systems},
  volume={35},
  pages={1966--1978},
  year={2022}
}

@inproceedings{qu2024goi,
  title={GOI: Find 3D Gaussians of Interest with an Optimizable Open-vocabulary Semantic-space Hyperplane},
  author={Qu, Yansong and Dai, Shaohui and Li, Xinyang and Lin, Jianghang and Cao, Liujuan and Zhang, Shengchuan and Ji, Rongrong},
  booktitle={Proceedings of the 32nd ACM International Conference on Multimedia},
  pages={5328--5337},
  year={2024}
}

@article{oquab2023dinov2,
  title={Dinov2: Learning robust visual features without supervision},
  author={Oquab, Maxime and Darcet, Timoth{\'e}e and Moutakanni, Th{\'e}o and Vo, Huy and Szafraniec, Marc and Khalidov, Vasil and Fernandez, Pierre and Haziza, Daniel and Massa, Francisco and El-Nouby, Alaaeldin and others},
  journal={arXiv preprint arXiv:2304.07193},
  year={2023}
}

@inproceedings{radford2021learning,
  title={Learning transferable visual models from natural language supervision},
  author={Radford, Alec and Kim, Jong Wook and Hallacy, Chris and Ramesh, Aditya and Goh, Gabriel and Agarwal, Sandhini and Sastry, Girish and Askell, Amanda and Mishkin, Pamela and Clark, Jack and others},
  booktitle={International conference on machine learning},
  pages={8748--8763},
  year={2021},
  organization={PMLR}
}

@inproceedings{zhai2023sigmoid,
  title={Sigmoid loss for language image pre-training},
  author={Zhai, Xiaohua and Mustafa, Basil and Kolesnikov, Alexander and Beyer, Lucas},
  booktitle={Proceedings of the IEEE/CVF International Conference on Computer Vision},
  pages={11975--11986},
  year={2023}
}

@article{liu2023grounding,
  title={Grounding DINO: Marrying DINO with Grounded Pre-Training for Open-Set Object Detection},
  author={Liu, Shilong and Zeng, Zhaoyang and Ren, Tianhe and Li, Feng and Zhang, Hao and Yang, Jie and Li, Chunyuan and Yang, Jianwei and Su, Hang and Zhu, Jun and Zhang, Lei},
  journal={arXiv preprint arXiv:2303.05499},
  year={2023}
}

@inproceedings{schonberger2016structure,
  title={Structure-from-motion revisited},
  author={Schonberger, Johannes L and Frahm, Jan-Michael},
  booktitle={Proceedings of the IEEE conference on computer vision and pattern recognition},
  pages={4104--4113},
  year={2016}
}

@inproceedings{schonberger2016pixelwise,
  title={Pixelwise view selection for unstructured multi-view stereo},
  author={Sch{\"o}nberger, Johannes L and Zheng, Enliang and Frahm, Jan-Michael and Pollefeys, Marc},
  booktitle={Computer Vision--ECCV 2016: 14th European Conference, Amsterdam, The Netherlands, October 11-14, 2016, Proceedings, Part III 14},
  pages={501--518},
  year={2016},
  organization={Springer}
}

@article{straub2019replica,
  title={The Replica dataset: A digital replica of indoor spaces},
  author={Straub, Julian and Whelan, Thomas and Ma, Lingni and Chen, Yufan and Wijmans, Erik and Green, Simon and Engel, Jakob J and Mur-Artal, Raul and Ren, Carl and Verma, Shobhit and others},
  journal={arXiv preprint arXiv:1906.05797},
  year={2019}
}

@inproceedings{yeshwanth2023scannet++,
  title={Scannet++: A high-fidelity dataset of 3d indoor scenes},
  author={Yeshwanth, Chandan and Liu, Yueh-Cheng and Nie{\ss}ner, Matthias and Dai, Angela},
  booktitle={Proceedings of the IEEE/CVF International Conference on Computer Vision},
  pages={12--22},
  year={2023}
}

@article{geiger2013vision,
  title={Vision meets robotics: The kitti dataset},
  author={Geiger, Andreas and Lenz, Philip and Stiller, Christoph and Urtasun, Raquel},
  journal={The International Journal of Robotics Research},
  volume={32},
  number={11},
  pages={1231--1237},
  year={2013},
  publisher={Sage Publications Sage UK: London, England}
}

@inproceedings{dai2017scannet,
  title={Scannet: Richly-annotated 3d reconstructions of indoor scenes},
  author={Dai, Angela and Chang, Angel X and Savva, Manolis and Halber, Maciej and Funkhouser, Thomas and Nie{\ss}ner, Matthias},
  booktitle={Proceedings of the IEEE conference on computer vision and pattern recognition},
  pages={5828--5839},
  year={2017}
}

@inproceedings{schops2017multi,
  title={A multi-view stereo benchmark with high-resolution images and multi-camera videos},
  author={Schops, Thomas and Schonberger, Johannes L and Galliani, Silvano and Sattler, Torsten and Schindler, Konrad and Pollefeys, Marc and Geiger, Andreas},
  booktitle={Proceedings of the IEEE conference on computer vision and pattern recognition},
  pages={3260--3269},
  year={2017}
}

@article{mobile_sam_v2,
  title={MobileSAMv2: Faster Segment Anything to Everything},
  author={Zhang, Chaoning and Han, Dongshen and Zheng, Sheng and Choi, Jinwoo and Kim, Tae-Ho and Hong, Choong Seon},
  journal={arXiv preprint arXiv:2312.09579},
  year={2023}
}

@article{aanaes2016large,
  title={Large-scale data for multiple-view stereopsis},
  author={Aan{\ae}s, Henrik and Jensen, Rasmus Ramsb{\o}l and Vogiatzis, George and Tola, Engin and Dahl, Anders Bjorholm},
  journal={International Journal of Computer Vision},
  volume={120},
  pages={153--168},
  year={2016},
  publisher={Springer}
}

@article{fan2024large,
  title={Large spatial model: End-to-end unposed images to semantic 3d},
  author={Fan, Zhiwen and Zhang, Jian and Cong, Wenyan and Wang, Peihao and Li, Renjie and Wen, Kairun and Zhou, Shijie and Kadambi, Achuta and Wang, Zhangyang and Xu, Danfei and others},
  journal={Advances in Neural Information Processing Systems},
  volume={37},
  pages={40212--40229},
  year={2024}
}

@inproceedings{engelmann2024opennerf,
  title     = {{OpenNeRF: Open Set 3D Neural Scene Segmentation with Pixel-Wise Features and Rendered Novel Views}},
  author    = {Engelmann, Francis and Manhardt, Fabian and Niemeyer, Michael and Tateno, Keisuke and Pollefeys, Marc and Tombari, Federico},
  booktitle = {International Conference on Learning Representations},
  year      = {2024}
}

@inproceedings{schroeppel2022robust,
  author     = {Philipp Schr\"oppel and Jan Bechtold and Artemij Amiranashvili and Thomas Brox},
  booktitle  = {Proceedings of the International Conference on {3D} Vision ({3DV})},
  title      = {A Benchmark and a Baseline for Robust Multi-view Depth Estimation},
  year       = {2022}
}

@article{duisterhof2024mast3r,
  title={MASt3R-SfM: a Fully-Integrated Solution for Unconstrained Structure-from-Motion},
  author={Duisterhof, Bardienus and Zust, Lojze and Weinzaepfel, Philippe and Leroy, Vincent and Cabon, Yohann and Revaud, Jerome},
  journal={arXiv preprint arXiv:2409.19152},
  year={2024}
}

@article{wang2025vggt,
  title={Vggt: Visual geometry grounded transformer},
  author={Wang, Jianyuan and Chen, Minghao and Karaev, Nikita and Vedaldi, Andrea and Rupprecht, Christian and Novotny, David},
  journal={arXiv preprint arXiv:2503.11651},
  year={2025}
}

@inproceedings{wang2024vggsfm,
  title={Vggsfm: Visual geometry grounded deep structure from motion},
  author={Wang, Jianyuan and Karaev, Nikita and Rupprecht, Christian and Novotny, David},
  booktitle={Proceedings of the IEEE/CVF conference on computer vision and pattern recognition},
  pages={21686--21697},
  year={2024}
}

@InProceedings{Yang_2025_Fast3R,
  title={Fast3R: Towards 3D Reconstruction of 1000+ Images in One Forward Pass},
  author={Jianing Yang and Alexander Sax and Kevin J. Liang and Mikael Henaff and Hao Tang and Ang Cao and Joyce Chai and Franziska Meier and Matt Feiszli},
  booktitle={Proceedings of the IEEE/CVF Conference on Computer Vision and Pattern Recognition (CVPR)},
  month={June},
  year={2025},
}

@article{wang20243d,
  title={3d reconstruction with spatial memory},
  author={Wang, Hengyi and Agapito, Lourdes},
  journal={arXiv preprint arXiv:2408.16061},
  year={2024}
}

@inproceedings{guo2025wildseg3d,
  title={Wildseg3d: Segment any 3d objects in the wild from 2d images},
  author={Guo, Yansong and Hu, Jie and Qu, Yansong and Cao, Liujuan},
  booktitle={Proceedings of the IEEE/CVF International Conference on Computer Vision},
  pages={5166--5176},
  year={2025}
}
}

\clearpage
\setcounter{page}{1}
\maketitlesupplementary

\section{Proof of Interpolation Properties}

This section provides the complete proofs for the two key properties of our area-weighted interpolation strategy, which underpin the effectiveness of the Pixel Embedding Disambiguation described in Sec.~3.3 of the main text.

\noindent\textbf{Proposition 1.} \textit{Vector Normalization:} The interpolated vector $\hat{\mathbf{F}}_B$ preserves unit norm, ensuring it remains within the original semantic embedding space.

\noindent\textit{Proof.} The norm of $\hat{\mathbf{F}}_B$ is given by:
\begin{equation}
\begin{split}
\label{eq3}
\|\hat{\mathbf{F}}_B\|^2 = \left\|  a\mathbf{F}_A + b \mathbf{F}_B \right\|^2.
\end{split}
\end{equation}
Expanding this expression:
\begin{equation}
\begin{split}
\label{eq4}
\|\hat{\mathbf{F}}_B\|^2
&=\left( a \mathbf{F}_A + b \mathbf{F}_B \right) \cdot \left( a \mathbf{F}_A + b \mathbf{F}_B \right) \\
&=a^2 \|\mathbf{F}_A\|^2 + 2 ab \mathbf{F}_A \cdot \mathbf{F}_B + b^2 \|\mathbf{F}_B\|^2.
\end{split}
\end{equation}
Since $\mathbf{F}_A$ and $\mathbf{F}_B$ are unit vectors:
\begin{equation}
\begin{split}
\label{eq4}
\|\mathbf{F}_A\| = 1, \|\mathbf{F}_B\| = 1, \mathbf{F}_A \cdot \mathbf{F}_B = \cos(\theta).
\end{split}
\end{equation}
Substituting these values, we get:
\begin{equation}
\begin{split}
\label{eq5}
\|\hat{\mathbf{F}}_B\|^2 &= \frac{1}{\sin^2(\theta)}(\sin^2((1-t)\theta)+ \sin^2(t\theta)\\
&+ 2 \sin((1-t)\theta) \sin(t\theta) \cos(\theta)).
\end{split}
\end{equation}
Using trigonometric identities:
\begin{equation}
\begin{split}
\label{eq6}
\sin^2(\theta)&=\sin^2((1-t)\theta)+ \sin^2(t\theta)\\
&+ 2 \sin((1-t)\theta) \sin(t\theta) \cos(\theta),
\end{split}
\end{equation}
we find that:
\begin{equation}
\begin{split}
\label{eq7}
\|\hat{\mathbf{F}}_B\|^2 = \sin^2(\theta) / \sin^2(\theta) = 1.
\end{split}
\end{equation}
Thus, $\hat{\mathbf{F}}_B$ is confirmed to be a unit vector. $\square$

\noindent\textbf{Proposition 2.} \textit{Semantic Guidance:} If $\mathbf{F}_A$ has a higher similarity to a reference semantic vector $\mathbf{F}_C$ than $\mathbf{F}_B$ does, then $\hat{\mathbf{F}}_B$ will also exhibit a higher similarity to $\mathbf{F}_C$ than $\mathbf{F}_B$ does. This steers the aggregated representation toward more semantically meaningful directions.

\noindent\textit{Proof.} The cosine similarity between $\hat{\mathbf{F}}_B$ and $\mathbf{F}_C$ is:
\begin{equation}
\begin{split}
\label{eq8}
\hat{\mathbf{F}}_B\cdot\mathbf{F}_C=a(\mathbf{F}_A\cdot\mathbf{F}_C)+b(\mathbf{F}_B\cdot\mathbf{F}_C).
\end{split}
\end{equation}
Since $\mathbf{F}_A\cdot\mathbf{F}_C>\mathbf{F}_B\cdot\mathbf{F}_C$, we have:
\begin{equation}
\begin{split}
\label{eq9}
a(\mathbf{F}_A\cdot\mathbf{F}_C)+b(\mathbf{F}_B\cdot\mathbf{F}_C)&>(a+b)(\mathbf{F}_B\cdot\mathbf{F}_C).
\end{split}
\end{equation}
Using trigonometric properties:
\begin{equation}
\begin{split}
\label{eq10}
a+b=\frac{\sin((1-t)\theta)}{\sin(\theta)}+\frac{\sin(t\theta)}{\sin(\theta)}=1,
\end{split}
\end{equation}
we conclude:
\begin{equation}
\begin{split}
\label{eq11}
\hat{\mathbf{F}}_B\cdot\mathbf{F}_C=a(\mathbf{F}_A\cdot\mathbf{F}_C)+b(\mathbf{F}_B\cdot\mathbf{F}_C)>\mathbf{F}_B\cdot\mathbf{F}_C.
\end{split}
\end{equation}
This confirms that $\hat{\mathbf{F}}_B$ semantically integrates information from both $\mathbf{F}_A$ and $\mathbf{F}_B$, steering the aggregated representation toward more meaningful directions. $\square$

\section{Extended Ablation Studies}

This section provides extended ablation studies that complement the analysis in Sec.~4.3 of the main paper, offering further insights into the design choices and parameter sensitivity of PE3R.

\subsection{Anomaly Point Selection}

The accuracy of our anomaly point detection mechanism is validated through both empirical and theoretical analysis.

\noindent\textbf{Empirical Validation.}  
We determine the anomaly threshold by statistically analyzing the mean 3D distance between spatially consistent points across semantic regions. This threshold effectively separates normal points from anomalies. Table~\ref{tab8} presents results under different thresholds on ScanNet, with the optimal performance observed at a threshold of 0.003.

\begin{table}[t]
\tablestyle{3.0pt}{1.0}
\centering
\begin{tabular}{c|cc}
\toprule
Threshold & rel$\downarrow$ & $\tau\uparrow$ \\
\midrule
0.001 & 6.1 & 49.3 \\
0.002 & 5.9 & 50.3 \\
0.003 & \textbf{5.5} & \textbf{55.1} \\
0.004 & 5.6 & 53.6 \\
0.005 & 5.8 & 52.8 \\
\bottomrule
\end{tabular}
\caption{Anomaly point selection with varying thresholds on ScanNet. The optimal value of 0.003 is used as the default in all main experiments.}\label{tab8}
\end{table}

\noindent\textbf{Theoretical Validation.}  
While anomaly detection could theoretically be approached through parametric modeling of intra-object distributions (\eg, using Gaussian estimation or least-squares fitting), such methods become computationally prohibitive in complex, large-scale scenes. Our empirical thresholding strategy provides a practical and scalable alternative that maintains high performance without introducing significant computational overhead.

\subsection{Sliding Window Size Analysis}

The size of the sliding window used for local semantic-aware distance computation significantly impacts both reconstruction quality and computational efficiency. As demonstrated in Table~\ref{tab9}, we evaluate three window sizes on the multi-view depth estimation task. A $5 \times 5$ window achieves the optimal trade-off, capturing sufficient local context for robust anomaly detection without introducing excessive computational cost or over-smoothing fine geometric details.

\begin{table}[t]
\tablestyle{3.0pt}{1.0}
\centering
\begin{tabular}{c|cc}
\toprule
Window Size & avg. rel$\downarrow$ & avg. $\tau\uparrow$ \\
\midrule
$3\times3$ & 4.9 & 65.5 \\
$5\times5$ & \textbf{4.5} & \textbf{68.0} \\
$7\times7$ & 4.8 & 66.1 \\
\bottomrule
\end{tabular}
\caption{Effect of sliding window size on anomaly detection performance across all depth estimation benchmarks.}\label{tab9}
\end{table}

\subsection{RGB Image Smoothing Analysis}

To enhance robustness in challenging regions characterized by reflections, transparency, or complex textures, we apply semantic-aware smoothing to the input images prior to pointmap estimation. The smoothing operation is defined as $\hat{\mathbf{y}} = \alpha \cdot \mathbf{x} + (1-\alpha) \cdot \mathbf{y}$, where $\mathbf{x}$ is the mean RGB value of the semantic region, $\mathbf{y}$ is the original pixel value, and $\alpha$ controls the smoothing strength.

As shown in Table~\ref{tab10}, moderate smoothing ($\alpha=0.1$) provides the optimal balance between noise suppression and detail preservation. Lower values ($\alpha \leq 0.1$) insufficiently address visual artifacts, while higher values ($\alpha \geq 0.2$) over-smooth genuine structural details, ultimately degrading geometric accuracy.

\begin{table}[t]
\tablestyle{3.0pt}{1.0}
\centering
\begin{tabular}{cc|cc}
\toprule
$a$ & $b$ & avg. rel$\downarrow$ & avg. $\tau\uparrow$ \\
\midrule
0.00 & 1.00 & 5.3 & 60.2 \\
0.10 & 0.90 & \textbf{4.5} & \textbf{68.0} \\
0.20 & 0.80 & 4.7 & 62.3 \\
0.50 & 0.50 & 4.9 & 61.6 \\
0.80 & 0.20 & 6.1 & 59.4 \\
0.90 & 0.10 & 6.5 & 57.5 \\
1.00 & 0.00 & 10.2 & 50.2 \\
\bottomrule
\end{tabular}
\caption{Effect of semantic-aware smoothing parameter $\alpha$ on reconstruction performance. Moderate smoothing ($\alpha=0.1$) achieves the best results.}\label{tab10}
\end{table}

\subsection{Iterative Refinement Analysis}

We further investigate the effect of repeated refinement iterations in the semantic point cloud reconstruction module. As shown in Table~\ref{tab11}, while a single iteration provides substantial improvement, additional iterations yield diminishing returns with increased computational cost. This suggests that our method achieves effective refinement in a single pass, maintaining the efficiency advantages of the overall feed-forward pipeline.

\begin{table}[t]
\tablestyle{3.0pt}{1.0}
\centering
\begin{tabular}{c|cccc}
\toprule
Iterations & 1 & 2 & 3 & 4 \\
\midrule
avg. rel$\downarrow$ & \textbf{4.5} & 4.6 & 4.6 & 4.6 \\
avg. $\tau\uparrow$ & \textbf{68.0} & 67.9 & 67.9 & 67.9 \\
\bottomrule
\end{tabular}
\caption{Effect of iterative refinement cycles on reconstruction performance (Mip-NeRF360). A single iteration achieves optimal performance.}\label{tab11}
\end{table}



\end{document}